\documentclass[12pt]{article}

\usepackage{setspace}
\usepackage[top=1in, bottom=1in, left=1in, right=1in]{geometry}
\usepackage{enumitem} 
\setlist{nolistsep} 

\usepackage{mathtools,amssymb,amsthm,amsthm,amsfonts,latexsym,bbm}
\usepackage{amscd,amsxtra} 
\usepackage{graphics,graphicx,xcolor}
\usepackage{ifpdf}
\usepackage[caption=false]{subfig}
\usepackage{multirow}
\usepackage[colorlinks,linkcolor=blue,citecolor=blue,urlcolor=blue]{hyperref}
\usepackage[comma]{natbib}
\usepackage{comment}
\usepackage{ulem}
\usepackage{authblk} 

\makeatletter
\renewcommand*{\@fnsymbol}[1]{\ensuremath{\ifcase#1\or \dagger\or \ddagger\or
   \mathsection\or \mathparagraph\or \|\or **\or \dagger\dagger
   \or \ddagger\ddagger \else\@ctrerr\fi}}
\makeatother

\allowdisplaybreaks
\theoremstyle{plain}
\newtheorem{theorem}{\sc Theorem}
\newtheorem{corollary}[theorem]{\sc Corollary}
\newtheorem{lemma}[theorem]{\sc Lemma}

\theoremstyle{remark}

\theoremstyle{definition}

\def\vv{\boldsymbol v}

\def\xv{\boldsymbol x}

\def\Xv{\boldsymbol X}

\newcommand{\Dc}{\mathcal{D}}

\newcommand{\Lc}{\mathcal{L}}

\newcommand{\Sc}{\mathcal{S}}

\newcommand{\Xc}{\mathcal{X}}
\newcommand{\Yc}{\mathcal{Y}}

\newcommand{\zetav}{\mbox{\boldmath$\zeta$}}
\newcommand{\etav}{\mbox{\boldmath{$\eta$}}}

\newcommand{\muv}{\mbox{\boldmath{$\mu$}}}

\newcommand{\xiv}{\mbox{\boldmath{$\xi$}}}

\newcommand{\omegav}{\mbox{\boldmath{$\omega$}}}
\newcommand{\Sigmav}{\mbox{\boldmath{$\Sigma$}}}

\newcommand{\thetav}{\mbox{\boldmath{$\theta$}}}

\def\1v{\mathbf 1}
\def\0v{\mathbf 0}
\def\Id{\mathbf I} 
\newcommand{\ind}[1]{\mathbbm{1}_{\left[ {#1} \right] }}
\newcommand{\Ind}[1]{\mathbbm{1}_{\left\{ {#1} \right\} }}

\newcommand{\R}{\mathbb R}

\newcommand{\E}{\mathbb E}
\newcommand{\sgn}{\mathop{\mathrm{sign}}}
\def\Pr{\mathbb P}

\newcommand{\diag}{\mathop{\rm diag}}

\newcommand{\ds}{\displaystyle}
\newcommand{\mb}{\mbox}
\newcommand{\wh}{\widehat}
\newcommand{\argmin}{\operatornamewithlimits{argmin}}

\newcommand{\set}[1]{\left\{#1\right\}}

\def\ie{\textit{i.e.}}

\title{Distance-weighted Support Vector Machine}
\author{Xingye Qiao\thanks{Corresponding author}}
\affil{Department of Mathematical Sciences\authorcr  State University of New York, Binghamton, NY 13902-6000.\authorcr E-mail: \texttt{qiao@math.binghamton.edu}}
\author{Lingsong Zhang}
\affil{Department of Statistics\authorcr Purdue University, West Lafayette, IN 47907.\authorcr E-mail: \texttt{lingsong@purdue.edu}}

\begin{document}

\maketitle 
\pagenumbering{Roman}

\centerline{\bf Abstract} {\small A novel linear classification method that possesses the merits of both the Support Vector Machine (SVM) and the Distance-weighted Discrimination (DWD) is proposed in this article. The proposed Distance-weighted Support Vector Machine method can be viewed as a hybrid of SVM and DWD that finds the classification direction by minimizing mainly the DWD loss, and determines the intercept term in the SVM manner. We show that our method inheres the merit of DWD, and hence, overcomes the data-piling and overfitting issue of SVM. On the other hand, the new method is not subject to imbalanced data issue which was a main advantage of SVM over DWD. It uses an unusual loss which combines the Hinge loss (of SVM) and the DWD loss through a trick of axillary hyperplane. Several theoretical properties, including Fisher consistency and asymptotic normality of the DWSVM solution are developed. We use some simulated examples to show that the new method can compete DWD and SVM on both classification performance and interpretability. A real data application further establishes the usefulness of our approach.

\vspace{0.15in} 
\noindent \textit{KEYWORDS}: Discriminant analysis; Fisher consistency; Imbalanced data; High-dimensional, low-sample size data; Support Vector Machine.

\newpage
\setlength{\belowdisplayskip}{0.5em} \setlength{\belowdisplayshortskip}{0.5em}
\setlength{\abovedisplayskip}{0.5em} \setlength{\abovedisplayshortskip}{0.5em}

\pagenumbering{arabic}
\setcounter{page}{1}
\section{Introduction}\label{sec:intro}
Classification is a very important research topic in statistical machine learning, and has many useful applications in various scientific and social research areas. In this article, we focus on the binary linear classification problem, in which a classification rule is to be found that maps a point in $\Xc$ to a class label chosen from $\Yc$, $\phi:~\Xc\mapsto\Yc$ where $\Xc=\R^d$ and $\Yc = \set{+1,-1}$. We focus on linear classification methods instead of nonlinear ones because they are easy to interpret due to simple formulations. In particular, each linear classification rule is associated with a linear discriminant function $f(\xv) = \xv^T\omegav+\beta$, where the coefficient direction vector $\omegav\in\R^d$ has unit $L_2$ norm, and $\beta\in\R$ is the intercept term. The classification rule is then $\phi(\xv)=\sgn(f(\xv))$, that is, the sample space $\R^d$ is divided into halves by the separating hyperplane defined by $\set{\xv:~f(\xv)\equiv \xv^T\omegav+\beta=0}$. The coefficient direction vector $\omegav$ determines the orientation of the hyperplane (as a matter of fact, it is the normal vector of this hyperplane), and the intercept term $\beta$ determines its location.

There is a large body of literature on linear classification. See \citet{Duda2001Pattern} and \citet{Hastie2009elements} for comprehensive introductions. Among many linear classification methods, the Support Vector Machine \citep[SVM;][]{Cortes1995Support, vapnik1998statistical, cristianini2000introduction} and the Distance-weighted Discrimination \citep[DWD;][]{marron2007distance, Qiao2010Weighteda} are two state-of-the-art instances and have received a lot of attention. A brief review of these two methods will be given in Section \ref{sec:svm_dwd}. 

In the high-dimensional, low-sample size (HDLSS) data setting, a so-called ``data-piling'' phenomenon has been observed for SVM \citep{marron2007distance} and some other classifiers \citep[for example,][]{Ahn2010maximal}. Data-piling is referred to the phenomenon that after projected to the direction vector $\omegav$ given by a linear classifier, a large portion of the data vectors pile upon each other and concentrate on two points. Data-piling reflects severe overfitting in the HDLSS data setting and is an indicator that the direction is driven by artifacts in the data, and hence the direction as well as the classification performance can be stochastically volatile. Moreover, it turns out that the directions from these linear classification methods are much deviated from the Bayes rule direction (when the Bayes rule exists and is linear). To this end, DWD was proposed largely to overcome the data-piling issue in the HDLSS setting and has been quite successful on that.

While DWD overcomes the data-piling and mitigates the overfitting effect, it is sensitive to the imbalanced sample sizes between the two classes \citep{Qiao2010Weighteda}. In particular, when the sample size of one class is much greater than the other one, the classification boundary would be pushed towards the minority class and consequently, all future data vectors will be classified into the majority class.

\begin{figure}[!ht]\vspace{-4ex}
	\centering
		\includegraphics[width=\linewidth]{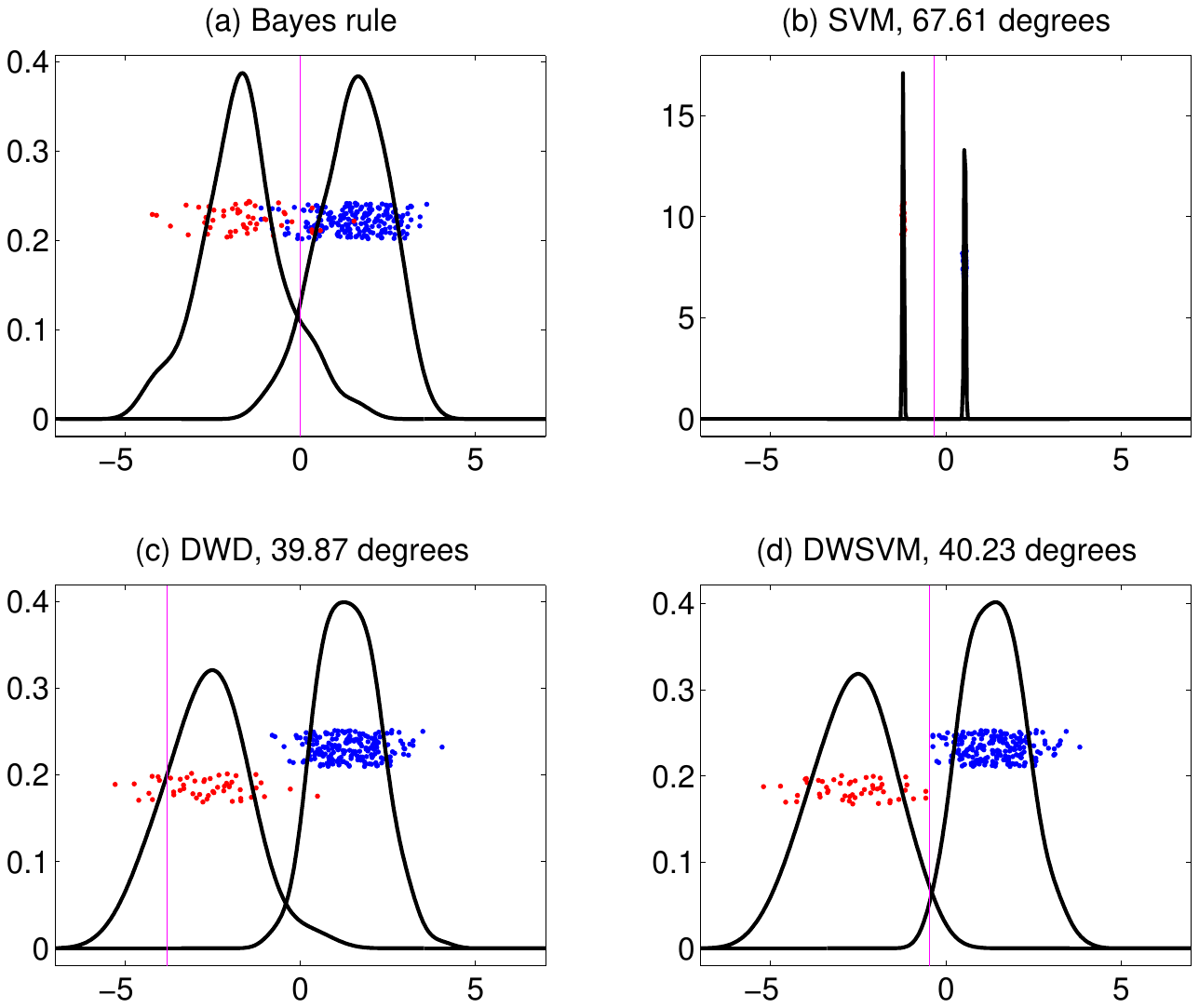}\vspace{-4ex}
	\caption{\small Plots of projections to: (a) the true mean difference (Bayes rule) direction, (b) the SVM direction, (c) the DWD direction and (d) the proposed DWSVM direction. The angles (in degree) between the last three directions and the first direction are shown in the titles. Projections of the separating hyperplanes of different methods are depicted by the magenta vertical lines. Panel (a) shows the Bayes direction and the separating hyperplane to be compared with. SVM in Panel (b) demonstrate very good separation between the two classes, but severe data-piling also appears. The projected data vectors are nowhere near Gaussian, which suggests that the direction is too much deviated from the Bayes direction in Panel (a). Panel (c) shows that DWD has no data-piling issue, and the projection plot preserves the Gaussian pattern. However, the separating hyperplane is pushed towards the red class because of its relatively small sample size. Our proposed DWSVM approach (Panel (d)) combines the merits of SVM and DWD. It preserves a good direction by showing the Gaussian pattern in the projections while finds a good intercept term which is not subject to imbalanced sample sizes.}
	\label{fig:ex1}
\end{figure}

\citet{Qiao13Flexible} have thoroughly studied the high-dimensional overfitting issue of SVM and the imbalanced data issue of DWD. Moreover, they proposed a new family of classifiers called FLAME which both SVM and DWD belong to. To illustrate the main points of data-piling and imbalanced issues, we show projection plots of a toy example to four different discriminant direction vectors in Figure \ref{fig:ex1}. In this example, the data vectors from the two classes are generated from multivariate normal distributions $N_d(\pm\mu\1v_d,\Id_d)$, where the dimension $d=300$, $\mu =1.35/\sqrt{d}=0.07794229$, $\1v_d$ is a $d$-dimensional vector of all ones and $\Id_d$ is the $d\times d$ identity matrix. The Bayes rule in this example has direction $\omegav_B = \1v_d/\sqrt{d}$ and the Bayes intercept $\beta_B=0$. Here the sample size of the positive class (with $Y=+1$) is $n_+=200$ and the negative class sample size is $n_-=50$.

Panel (a) in Figure \ref{fig:ex1} shows the true mean difference direction (which in fact is the Bayes direction) and the projections of the data vectors therein. They serve as the benchmark to be compared with. Panel (b) is for the SVM direction and it demonstrates a very dramatic separation between the two classes. This could be an alarming bell for overfitting. Indeed, severe data-piling is visible. The projected data vectors are nowhere near Gaussian, which suggests that the direction is too much deviated from the true direction in Panel (a). This deviation is also measured by the angle between the SVM direction and the Bayes direction (67 degrees, shown in the title). Panel (c) shows that DWD has no data-piling issue, and the projection plot preserves the Gaussian pattern, which means that there is some potential to interpret the data using the DWD direction. However, because the blue class (positive class with $Y=+1$) has four times sample size as the red class, the separating hyperplane is therefore pushed towards the red class. Expectedly, its classification performance is not good.

In this article, we propose a new method which integrates the merits of SVM and DWD, and thus can address the data-piling issue and the imbalanced data issue at the same time. Our proposed method is named Distance-weighted Support Vector Machine (DWSVM) to salute the above two classical methods. As shown in Panel (d) of Figure \ref{fig:ex1}, DWSVM preserves a good direction by showing the Gaussian pattern in the projections while finds a good intercept term which is not subject to the imbalanced sample sizes.  In addition, we prove in theory that the DWSVM is Fisher consistent and asymptotically normal, and that its intercept term is not sensitive to imbalanced sample size as DWD is.

The rest of the article is organized as follows. Section \ref{sec:svm_dwd} gives a brief introduction to the SVM and the DWD methods. Our DWSVM method is proposed in Section \ref{sec:DWSVM}. Simulated examples and a real application are studied in Sections \ref{sec:simulation} and \ref{sec:real}. Several theoretical results are given in Section \ref{sec:theory}. Some concluding remarks are made in Section \ref{sec:conclude}.

\section{Classical methods}\label{sec:svm_dwd}
In this section, we give a brief introduction to SVM and DWD, their formulations and the discussion on the roles of different terms.
\subsection{Classification and Loss Functions}
In classification, one is given a training data set, $\Dc\equiv \set{(\xv_i,y_i)\in\Xc\otimes\Yc,i=1,\dots,n}$ and the goal is to find a rule, $\phi(\xv)\equiv\sgn(f(\xv))$, depending on $\Dc$,  so that the classification error $\E(\phi(\Xv)\neq Y)$ is minimized. A natural estimate of the classification error is $\frac{1}{n}\sum_{i=1}^n \ind{\sgn(f(\xv_i))\neq y_i}=\frac{1}{n}\sum_{i=1}^n \ind{y_if(\xv_i)< 0}$. However, even in the simple case of linear classification where $f(\xv)$ is assume to have the form $f(\xv)=\xv^T\omegav+\beta$, searching for $(\omegav,\beta)$ to minimize $\sum_{i=1}^n \ind{y_if(\xv_i)< 0}$ is intractable due to the discontinuity and nonconvexity of the objective function. In statistical learning, a common practice to avoid these issues is to use a convex surrogate function to approximate/upper-bound the 0-1 loss function $\ind{yf(\xv)<0}$. For any discriminant function $f(\xv)$, let us define $u\equiv yf(\xv)$ the functional margin which can be viewed as the signed distance (up to a constant) from data point $\xv$ to the separating hyperplane $\set{\xv:f(\xv)=0}$. A convex surrogate $\psi(u): \R\mapsto \R^+$ can be used in the place of $\ind{u<0}$. For example, a classification rule can be obtain by,
\begin{align*}
\min_{\omegav,\beta} \sum_{i=1}^n \psi\left(y_i(\xv_i^T\omegav+\beta)\right)+\frac{\lambda}{2}\|\omegav\|^2
\end{align*}
Here, the first term in the objective function bounds the empirical classification error and the  $\|\omegav\|^2$ term in the second term measures the complexity of the model. The choice of the tuning parameter $\lambda$ balances the two main concerns. Equivalently, this optimization problem can be cast to
$\min_{\omegav,\beta} \sum_{i=1}^n \psi\left(y_i(\xv_i^T\omegav+\beta)\right),~~\mb{s.t.}~\|\omegav\|^2\leq C$ due to standard optimization theory.
Many classification methods fall into this category, such as Support Vector Machine, AdaBoost \citep{Freund1997}, and logistic regression \citep{Friedman2000}. See \citet{Bartlett2006Convexity} and the references therein for more sophisticated discussion on convex loss functions and their implications for risk bounds.

\subsection{Support Vector Machine (SVM)}
By choosing the hinge loss function $(1-u)_+$ as the convex surrogate, where $\left(a\right)_+\equiv \max(a,0)$ is the positive part of $a$, the SVM method is defined to maximize the smallest distances of all observations to the separating hyperplane. Mathematically, for some positive $\lambda$, the optimization problem of SVM can be written as $\ds\min_{\tilde\omegav,\tilde\beta}~\sum_{i=1}^n \left(1-y_i  (\xv_i^T\tilde\omegav+\tilde\beta)\right)_++\frac{\lambda}{2}\|\tilde\omegav\|^2$. Here, in addition to measuring the model complexity, $\|\omegav\|^2$ also defines a notion of gap between the two classes for SVM. In particular, $2/\|\tilde\omegav\|$ is the distance between the classes (up to a constant). Hence, to minimize $\|\tilde\omegav\|^2$ is the same as to maximize the gap between classes. The notion of gap will play a central role in the derivation of methods in this article.

The formulation above can be equivalently written as $\ds{\min_{\tilde\omegav,\tilde\beta}~\sum_{i=1}^n \left(1-y_i  (\xv_i^T\tilde\omegav+\tilde\beta)\right)_+, ~\mb{s.t. } \|\tilde\omegav\|^2\leq C}$. Here the coefficient vector $\tilde\omegav$ does not necessarily have unit norm. We let $\omegav = \tilde\omegav/\sqrt{C}$ and $\beta = \tilde\beta/\sqrt{C}$. Then the SVM solution is given by
$\ds\argmin_{\omegav,\beta}~\sum_{i=1}^n \left(\sqrt{C}-Cy_i (\xv_i^T\omegav+\beta)\right)_+$, $\mb{s.t. } \|\omegav\|^2\leq 1$. In this formulation, a modified Hinge loss function,
\begin{align}\label{modifiedHinge}
 H_C(u)=\left\{
	     \begin{array}{cc}
              \sqrt{C}-Cu & \mb{if } u\leq \frac{1}{\sqrt{C}},\\
	      0 & \mb{otherwise},
             \end{array}\right.
\end{align} is used, such that SVM can be viewed as to minimize $\sum_{i=1}^n H_C(u_i)$, subject to $\|\omegav\|^2\leq 1$, where the functional margin $u_i$ for the $i$th data is $u_i=y_i (\xv_i^T\omegav+\beta)$.  In order to align this formulation with that of DWD, we introduce a slack variable $\xi_i$ and rewrite SVM as,
\begin{align}
	\argmin_{\omegav,\beta,\xi_i}\quad & \sum_{i=1}^n \xi_i~,\label{eq:svm_formula1}\\
	\mb{s.t. } & Cy_i (\xv_i^T\omegav+\beta)+\xi_i\geq \sqrt{C},~\xi_i\geq 0,\label{eq:svm_formula2}\\
	\quad &\|\omegav\|^2\leq 1.\label{eq:svm_formula3}
\end{align}

\subsection{Distance-weighted Discrimination (DWD)}
DWD method was proposed by \citet{marron2007distance} to improve the performance of SVM in the HDLSS setting. It also maximizes a notion of gap between classes: the harmonic mean of the distances of all data vectors to the separating hyperplane. Let $r_i=y_i(\xv_i^T\omegav+\beta)+\eta_i$ be the (adjusted) distance of the $i$th  data vector to the separating hyperplane. Mathematically, the solution of DWD is
\begin{align}
\argmin_{\omegav,\beta,\eta_i} \quad &\sum_{i=1}^n \left(\frac{1}{r_i}+C\eta_i\right)~,\label{eq:dwd_formula1}\\
\mb{s.t.} \quad& r_i=y_i(\xv_i^T\omegav+\beta)+\eta_i,~r_i\geq 0\mb{ and }\eta_i\geq 0,\label{eq:dwd_formula2}\\
 &\|\omegav\|^2\leq 1.\label{eq:dwd_formula3}
\end{align}
When $\eta_i=0$ and $y_i(\xv_i^T\omegav+\beta)>0$, $r_i=y_i(\xv_i^T\omegav+\beta)$ is the positive distance from each data vector to the separating hyperplane, due to (\ref{eq:dwd_formula2}). Thus $\sum_{i=1}^n 1/r_i$ defines a different notion of gap between classes from that by SVM (which was $2/\|\omegav\|$.)

If a positive distance $y_i(\xv_i^T\omegav+\beta)$ is not achievable for a data vector, then a positive slack variable $\eta_i$ is added to make $r_i$ positive.  Note that the value of correction $\eta_i$ corresponds to the amount of misclassification for the $i$th vector, and hence in order to minimize the misclassification, we must control $\sum_{i=1}^n\eta_i$ in the objective function.

We will use this formulation and combine it with that of the SVM method in (\ref{eq:svm_formula1})--(\ref{eq:svm_formula3}). Here, in order to understand the underlying DWD loss function for later use, we modify (\ref{eq:dwd_formula1})--(\ref{eq:dwd_formula3}) as follows. For each $i$, the term in the objective function $\left(\frac{1}{r_i}+C\eta_i\right)$ can be minimized over $\eta_i$. Some algebraic manipulations reveal that the optimization problem (about $\omegav$ and $\beta$) becomes 
\begin{align}
\argmin_{\omegav,\beta} \quad &\sum_{i=1}^n V_C\left(y_i(\xv_i^T\omegav+\beta)\right)~,\\
\mb{s.t.} \quad& \|\omegav\|^2\leq 1,
\end{align}
where the DWD loss function is defined as 
\begin{align}\label{dwdloss}
 V_C(u)=\left\{
	     \begin{array}{cc}
              2\sqrt{C}-Cu & \mb{if } u\leq \frac{1}{\sqrt{C}},\\
	      			1/u& \mb{otherwise}.
       \end{array}\right.
\end{align}

One key observation is to be made here. There are two main tasks in a binary linear classification method: 
\begin{enumerate}
\item a notion of gap which is to be maximized so as to make the two classes more separated; and 
\item a measure of misclassification which is to be minimized.
\end{enumerate}
Recall that in the SVM formulation, the notion of gap is $2/\|\omegav\|$, and the misclassification is measured by the Hinge loss function. SVM jointly minimizes the sum of these two components to search for a solution. In contrast, the DWD loss function in (\ref{dwdloss}) (derived from the objective function (\ref{eq:dwd_formula1})) has two functionalities: the first term $\sum_ir_i^{-1}$ in (\ref{eq:dwd_formula1}), the sum of inverse distance, is a notion of gap, and the second term $\sum_{i=1}^n\eta_i$ in (\ref{eq:dwd_formula1}) measures misclassification. The constraint $\|\omegav\|^2\leq 1$ in (\ref{eq:dwd_formula3}) merely serves as a regulator but it does not maximize the gap or minimize the misclassification. This appears to be a reason that DWD fails to provide a sensible intercept term for classification cutoff point: it cannot accomplish both tasks at the same time! 

The main motivation of our DWSVM approach is to extract the role of misclassification controller from the DWD loss, and assign this role to a SVM component. As will be shown in the next section, we carefully design our formulation to allow a DWD component to define a notion of gap between the two classes, which helps to find a good direction vector. Meanwhile, we let an SVM component to control the misclassification, which helps to search for a better intercept term.

\section{Distance-weighted Support Vector Machine}\label{sec:DWSVM}
In Section \ref{sec:alternative}, we first introduce a method which can be intuitively viewed as the prototype of the hybridization between SVM and DWD. Our proposed main method will be discussed in Section \ref{sec:DWSVM-sub1}. Some explanations to our method are given in Section \ref{sec:exp}.

\subsection{Simple prototype: naive DWSVM}\label{sec:alternative}
Before we introduce the DWSVM method, we discuss an intuitive hybridization between SVM and DWD, which is called the naive DWSWD method (nDWSVM). Based on the previous discussion and other results in the literature, a linear classifier with a direction given by DWD and an intercept term found by SVM is desirable. However, naively matching a DWD direction and an SVM intercept together would be problematic because the intercept would lose its context without the corresponding discriminant direction. Instead, we could train a DWD classifier on the data set, discard the DWD intercept, keep the DWD direction, and project all the data vectors to the 1-dimensional DWD direction to obtain a set of 1-dimensional data points. Lastly, find an intercept (a cutoff) by applying SVM to this 1-dimensional data set. Following this paradigm, we can get a DWD direction, which is thought to be better than an SVM direction in overcoming overfitting, and then given this DWD direction, search for an intercept in an SVM manner so as to mitigate the imbalanced data issue. We name this two-step procedure as nDWSVM. The nDWSVM method is a simple prototype of DWSVM, where the DWD component and the SVM component are trained separately.

\subsection{DWSVM}\label{sec:DWSVM-sub1}
In this subsection, we formally define the Distance-weighted Support Vector Machine (DWSVM) in order to improve nDWSVM. DWSVM simultaneously minimizes both the SVM loss function and the DWD loss function, to identify a common discriminant direction. The less-imbalance-sensitive SVM-driven intercept term will be used to identify the location of the optimal separating hyperplane. Mathematically, the optimization problem can be written as follows: Let $C_{dwd}>0$, $C_{svm}>0$ and $\alpha\in[0,1)$. The DWSVM classifier is given by the following optimization problem.
\begin{align}
\argmin_{\framebox{$\omegav,\beta$}\beta_0,\xi_i,\eta_i} \quad &\sum_{i=1}^n \left\{\alpha\left(\frac{1}{r_i}+C_{dwd}\cdot \eta_i\right)+(1-\alpha)\xi_i\right\},\label{eq:dwsvm_formula1}\\
\mb{s.t.}\quad 	& r_i=y_i(\xv_i^T\omegav+\beta_0)+\eta_i,~r_i\geq 0 \mb{ and } \eta_i\geq 0,\label{eq:dwsvm_formula2}\\
								& C_{svm}y_i (\xv_i^T\omegav+\beta)+\xi_i\geq \sqrt{C_{svm}},~\xi_i\geq 0,\label{eq:dwsvm_formula3}\\
								&\|\omegav\|^2\leq 1.\label{eq:dwsvm_formula4}
\end{align}
Importantly, in the end, we let $f(\xv)\equiv \xv^T\omegav+\beta$ and use $\sgn(f(\xv)) = \sgn(\xv^T\omegav+\beta)$ as the classification rule instead of $\sgn(\xv^T\omegav+\beta_0)$. Thus $\omegav$ and $\beta$ are the only two variables that really participate in classifying future data vectors, while $\beta_0$ is not involved. However, it does not mean that $\beta_0$ is of no significance. We will elaborate this point later.

Comparing (\ref{eq:dwsvm_formula1})--(\ref{eq:dwsvm_formula4}) with (\ref{eq:svm_formula1})--(\ref{eq:svm_formula3}) and (\ref{eq:dwd_formula1})--(\ref{eq:dwd_formula3}), we can see that the first term in (\ref{eq:dwsvm_formula1}) and the constraint (\ref{eq:dwsvm_formula2}) are similar to (\ref{eq:dwd_formula1}) and (\ref{eq:dwd_formula2}), while the second term in 
(\ref{eq:dwsvm_formula1}) and  the constraint (\ref{eq:dwsvm_formula3}) are similar to (\ref{eq:svm_formula1}) and (\ref{eq:svm_formula2}). Thus we can write the DWSVM formulation (\ref{eq:dwsvm_formula1})--(\ref{eq:dwsvm_formula4}) as \begin{align}
\argmin_{\omegav,\beta,\beta_0} \quad &\sum_{i=1}^n \left\{\alpha V_{C_{dwd}}(y_i(\xv_i^T\omegav+\beta_0))+(1-\alpha)H_{C_{svm}}(y_i (\xv_i^T\omegav+\beta))\right\},\label{eq:dwsvm_simple1}\\
\mb{s.t.}\quad 	&\|\omegav\|^2\leq 1.\label{eq:dwsvm_simple2}
\end{align}One might think that our DWSWM is just an optimization problem with the objective function equaling to a weighted average of the DWD loss and the SVM loss. However, it is more sophisticated than that. In the next subsection, we give some explanations to different components and parameters in DWSVM to help understand the new method.

\subsection{Understanding DWSVM}\label{sec:exp}
\begin{figure}[!t]\vspace{-10ex}
	\centering
		\includegraphics[width=\linewidth]{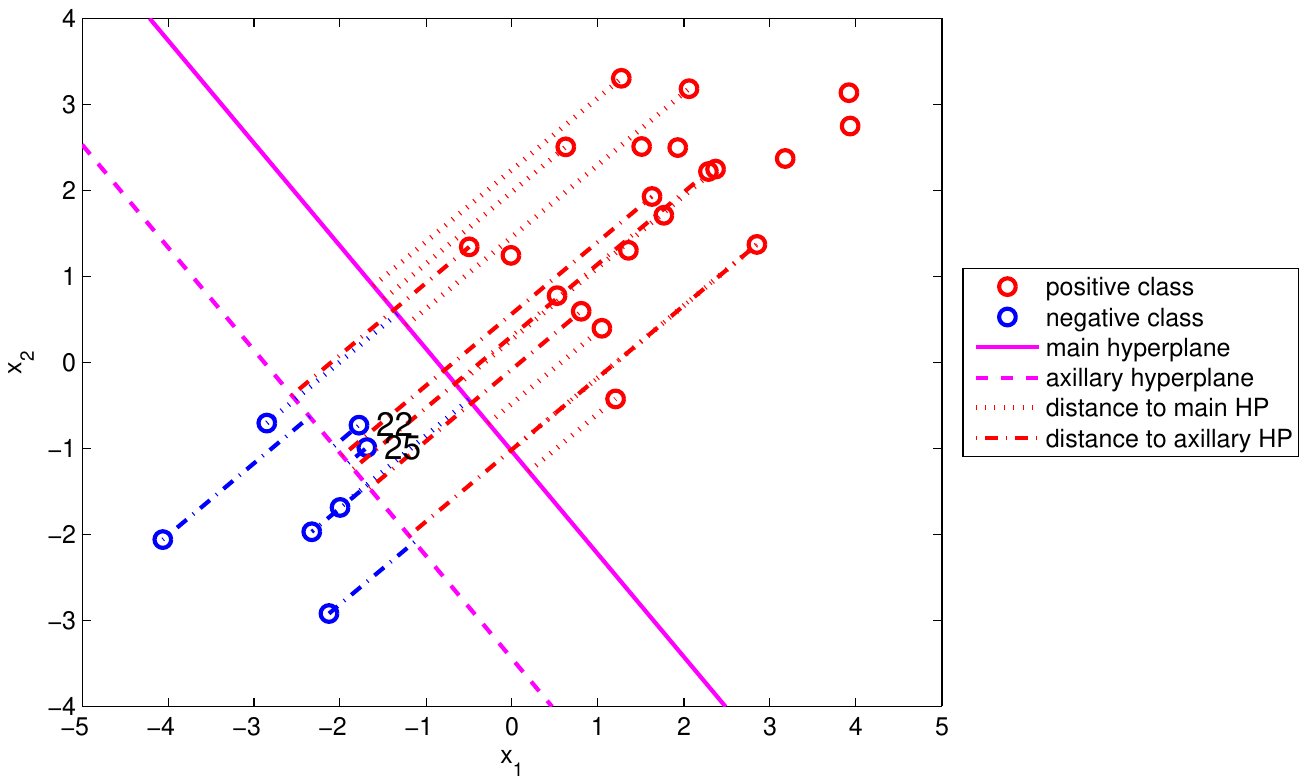}\vspace{-10ex}
	\caption{\small The main separating hyperplane (magenta solid line) and the axillary hyperplane (magenta dashed line) for DWSVM applied to a two-dimensional toy example. The distance from each data vector to the main hyperplane is depicted as a dotted line segment while the distance to the axillary hyperplane is depicted as a dotted-dashed line segment. Although the data vectors \#22 and \#25 are on the wrong side of the axillary hyperplane, they are not treated as misclassified by this method as they are both on the correct side of the main hyperplane. A positive $\eta_i$ is added to each negative functional margin $y_i(\xv_i^T\omegav+\beta_0),~i=22,25$, to make the sum positive.}
	\label{fig:fig2}
\end{figure}

\subsubsection*{Two hyperplanes}
First of all, there are two intercept terms $\beta_0$ and $\beta$ and only one direction vector $\omegav$ in the DWSVM method, that is, there are two hyperplanes that are parallel to each other, $\set{\xv:~\xv^T\omegav+\beta=0}$ and $\set{\xv:~\xv^T\omegav+\beta_0=0}$. For convenience, we call them the main hyperplane and the axillary hyperplane, respectively, and their corresponding discriminant functions $f\equiv \xv^T\omegav+\beta$ and $f_0\equiv \xv^T\omegav+\beta_0$ . See Figure \ref{fig:fig2} for an illustration using a two-dimensional toy example. In the plot, the magenta solid line is the main hyperplane and the magenta dashed line is the axillary hyperplane. 

\subsubsection*{Axillary hyperplane}
Note that $f_0$ is involved with the definition of $r_i$, the adjusted distance of a data vector to the axillary hyperplane, shown as dot-dashed line segments in Figure \ref{fig:fig2}. Similar to its role in DWD, $\sum_{i=1}^n \left(1/r_i\right)$ controls the gap between the two classes. In particular, the smaller $\sum_{i=1}^n \left(1/r_i\right)$ is, the more separated the two classes are.

In words, the purpose of the axillary hyperplane is not for classifying data vectors, but to make it possible to define a number of distances (from data vectors to itself) so that we can minimize the sum of the inverse distances. In the ordinary DWD, this axillary hyperplane has to coincide with the hyperplane that is actually used for classification. But here we allow some flexibility so that it is free of such restriction.

\subsubsection*{Necessity of the slack variable $\eta_i$}
When $y_if_0(\xv_i)<0$, the (signed) distance from the data vector to the axillary hyperplane is negative. In this case, a positive $\eta_i$ is added to $y_if_0(\xv_i)$ to make their sum $r_i$ positive. For example, in Figure \ref{fig:fig2}, the data vectors \#22 and \#25 are on the wrong side of the axillary hyperplane, hence both functional margins, $y_if_0(\xv_i),~i=22,25$, are negative. The DWSVM optimization adds some positive $\eta_i$'s to make the sum $r_i=y_if_0(\xv_i)+\eta_i$ positive. It is the sum of the inverse of $r_i$ that we minimize, instead of the sum of inverse of the signed distances $y_if_0(\xv_i)$. This adjustment is necessary. Otherwise, one can always make $\beta_0$ to be infinity, \ie, the axillary hyperplane is infinitely far from the data so that all the distances $y_if_0(\xv_i)$'s are infinity (positive or negative), and hence $1/(y_if_0(\xv_i)) = 0$. This is certainly not a desired situation because it would make the direction vector trivial (because the minimal of the objective function would always be 0 regardless of the choice of the direction). For these reasons, the addition of $\eta_i$ and the inclusion of $\sum_{i=1}^n \eta_i$ in the objective function are necessary to make the optimization problem meaningful.

\subsubsection*{Slack variable $\eta_i$ does not measure misclassification}
In the original DWD, the reason to minimize $\sum_{i=1}^n \eta_i$ is to control misclassification. However, the slack variable $\eta_i$ here is with respect to the axillary hyperplane (which is not useful in classification), rather than to the main separating hyperplane. Thus, we have liberated the DWD component from the burden of controlling misclassification, so that it can focus on defining the notion of gap and help searching for an optimal direction vector in the DWD fashion which overcomes overfitting.

\subsubsection*{Slack variable $\xi_i$}
Last of all, the second term $\xi_i$ in (\ref{eq:dwsvm_formula1}) is a proxy of the modified Hinge loss function of SVM in (\ref{modifiedHinge}). Inclusion of this term is for the purpose of controlling misclassification, because $\xi_i$ can be seen as $\left(\sqrt{C_{svm}}-C_{svm}u_i\right)_+$ where $u_i$ is the functional margin $y_i f(\xv_i)$ with respect to the main  hyperplane. Minimizing the sum of $\xi_i$'s can help to increase the functional margin $u_i$'s. Note that the functional margin $u_i=y_i f(\xv_i)$ can be interpreted as the distance to the main hyperplane (instead of the axillary one), which is ultimately the hyperplane that is used for classifying new data.

\subsubsection*{Summary}
In summary, the hyperplane defined by $\omegav$ and $\beta_0$ is an axillary hyperplane which is useful for finding the \textit{best} direction, and the one defined by $\omegav$ and $\beta$ is the main hyperplane that is useful for search the intercept and for good classification performance. By the trick of allowing two intercept terms, we gain some flexibility and manage to get two hyperplanes to each do their own job.

Empirically, nDWSVM can be used to approximate DWSVM, especially for low to moderate dimensions. Moreover, nDWSVM is very easy to implement, so long as the user has accessible implementations for both SVM and DWD (both are now available in R and MATLAB). The differences between DWSVM and nDWSVM are that in the two-step prototype nDWSVM, the direction is determined \textit{only} by the DWD algorithm, and the intercept is found by SVM based on the projections given by the DWD direction. However, in DWSVM, the axillary hyperplane (concerning DWD) and the main hyperplane (concerning SVM) work together to find the optimal direction. The optimization is done all at once in DWSVM.

Between DWD and DWSVM, the latter inherits the direction of the former, and adopts a very effective intercept term from its SVM component. Compared with SVM, the DWSVM method has a direction that is much improved due to the DWD component.

\section{Simulations}\label{sec:simulation}
In this section, we first compare the classification and the interpretability performance between the DWSVM approaches and the original SVM and DWD. The classification performance is measured by the misclassification rate for a large test data set with 4000 observations. The interpretability is a concept that is more of less vague. We partially measure it by the angle between the discriminant direction vector for the classifier under investigation and for the Bayes classifier. We believe the closer to the Bayes rule direction, the better the interpretability of the linear classifier is.

\subsection{Performance comparison}
We consider two different simulation settings. In each setting, samples from the two classes are generated from multivariate normal distributions $N_d(\pm\muv,\Sigmav)$.
\begin{enumerate}
		\item \textbf{Example 1}: Constant mean difference, identity covariance matrix example. $\muv \equiv c\1v_d$, and $\Sigmav \equiv \Id_d$, where $c>0$ is a scaling factor which makes $2c\|\1v_d\|_2 = 2.7$. This corresponds to the Mahalanobis distance between the two classes and represents a reasonable difficulty of classification using the Bayes rule.
		\item \textbf{Example 2}: Decreasing mean difference, block-diagonal interchangeable covariance matrix example. Here we let $\muv \equiv c\vv_d$, where $\vv_d = (\sqrt{50},\sqrt{49},\dots, \sqrt 1,0,0,\dots,0)^T\in\R^d$, and $\Sigmav \equiv \mb{Block-Diag}\set{\Sigma,\Sigma,\dots,\Sigma}$, where each $\Sigma$ is an $50\times 50$ interchangeable sub-covariance matrix whose diagonal entries are all 1 and off-diagonal entries are 0.8. The scaling factor $c$ is chosen to make the Mahalanobis distance $\left\{(2c\vv_d)\Sigmav^{-1}(2c\vv_d)^T\right\}^{1/2}=2.7$.
\end{enumerate}
In both simulation settings, we let the positive class sample size be 200 and the negative class sample size be 50. We vary the dimensions $d$ among $100,200,300,500$ and $1000$, thus the last three cases correspond to the HDLSS data settings.

\begin{figure}[!ht]\vspace{-2ex}
	\centering
		\includegraphics[width=0.8\linewidth]{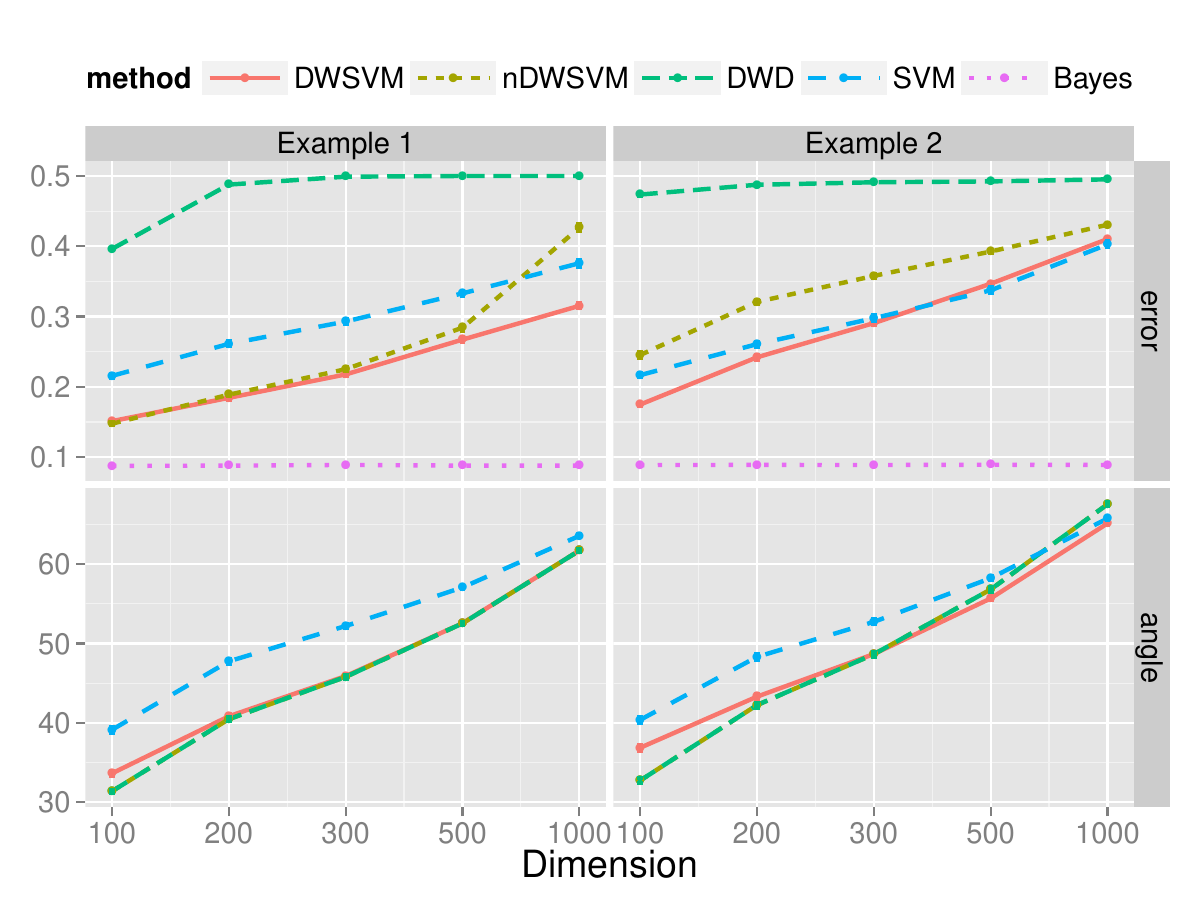}\vspace{-2ex}
	\caption{\small Comparison between four methods for Example 1 (the left panel) and Example 2 (the right panel). The misclassification error rates are shown on the top row and the angles between the classification directions and the Bayes direction are shown on the bottom. For Example 1 (left), for smaller dimensions, the two DWSVM approaches are better than SVM and DWD in terms of classification. For a large dimension, SVM outperforms the nDWSVM approach. The one-step DWSVM approach dominates all the other approaches in terms of classification performance. In terms of the interpretability (bottom) in Example 1, the two DWSVM approaches and the DWD approach all give similar and better results than the SVM approach. For Example 2 (right), DWSVM has similar good classification performance to SVM and similar good interpretability performance to DWD and nDWSVM. For small and moderate dimensions, the classification performance of DWSVM is significantly better than SVM.}
	\label{fig:bothsimulations}
\end{figure}

\subsubsection{Example 1}
In the top-left panel of Figure \ref{fig:bothsimulations}, we report the misclassification error of DWSVM, nDWSVM, DWD and SVM applied to a test data set with 2000 data points in each class which are generated according to the Constant mean difference, identity covariance matrix example. We conduct the simulation for 100 times and report the averages of the measurements. Our DWSVM approach uniformly gives the best classification results. The two-step alternative nDWSVM has very similar performance for dimensions 100, 200 and 500, but its performance is downgraded for higher dimensions. For all dimensions, unsurprisingly, the original DWD has misclassification rate close to almost 50\%, which is largely due to its intercept term which is subject to the imbalanced data.

In the bottom-left panel of Figure \ref{fig:bothsimulations}, we calculate the angles between the directions from different classifiers and the Bayes direction (for both simulation settings in this article, the Bayes classifiers are linear and the Bayes directions are well defined.) It shows that all the DWD related classifiers give very similar angles. As a matter of fact, the angles from DWSVM, DWSVM and DWD almost overlap with each other in this plot, except for low dimensional case where the DWSVM angle is a bit larger than the other two. On the other hand, the SVM directions are significantly more different from the Bayes direction than the DWD family directions are.

The observations so far verify the conjecture that DWD is worse at misclassification rate and SVM is worse at giving interpretable classification direction. DWSVM and nDWSVM appear to be able to address both issues simultaneously.

In the simulations, we tune the parameter $C_{svm}$ for SVM from a grid of possible values $2^{-5},2^{-4}\dots,2^{11},2^{12}$ and choose the one which gives rise to the smaller misclassification rate for a tuning data set that is identical to the training data set in terms of sample size and underlying distributions. For the DWD family of classifiers (DWSVM, nDWSVM and DWD), we let $C_{dwd}$ be 100 divided by a scaling factor that counts for the scale of the data, which was recommended by \citet{marron2007distance}. We fix $C_{svm}=100$ for DWSVM and nDWSVM. Lastly, we let $\alpha=0.5$ for DWSVM in our simulation study. Thus, the tuning parameter for SVM has been optimized while tuning parameters for our DWSVM methods are not tuned. Yet, our DWSVM method can achieve the performance as good as, sometimes even much better than, the other methods, for multiple criteria (classification and interpretability). This suggests a great potential of the DWSVM method.
\subsubsection{Example 2}
We have conducted the same comparison for the Decreasing mean difference, block-diagonal interchangeable covariance matrix example (Example 2) and the results are shown in the right panel of Figure \ref{fig:bothsimulations}. This time, the classification performance of DWSVM and SVM are closely competing with each other. For dimensions $d=100,200,300$, the DWSVM misclassification rates are smaller than SVM. But for dimensions $d=500$ and 1000, its classification error rates are slightly greater than SVM (not statistically significant). In terms of the angles between the classification direction vectors and the Bayes direction, the DWSVM direction are similar to those from nDWSVM and DWD, while all three are better than SVM. For the highest dimension case, all four directions are much different from the Bayes direction. However, the DWSVM direction is the best in this situation.

\subsection{Sensitivity to parameter values}
In this subsection, we study the impacts of different parameter values to DWSVM. First, we use ordinary DWD to search for an optimal choice of the $C_{dwd}$ parameter and fix its value in the sequel. In particular, we adopt the recommendation of $C_{dwd}$ in \citet{marron2007distance}. We choose not to further pursue in the direction of $C_{dwd}$ because this parameter has been well studied for ordinary DWD by \citet{marron2007distance} and for weighted DWD by \citet{Qiao2010Weighteda}. Here, we use simulation to illustrate the sensitivity of the DWSVM method to the difference choices of the other two parameters, $C_{svm}$ and $\alpha$.

\begin{figure}[!hb]\vspace{-5ex}
	\centering
		\includegraphics[width=\linewidth]{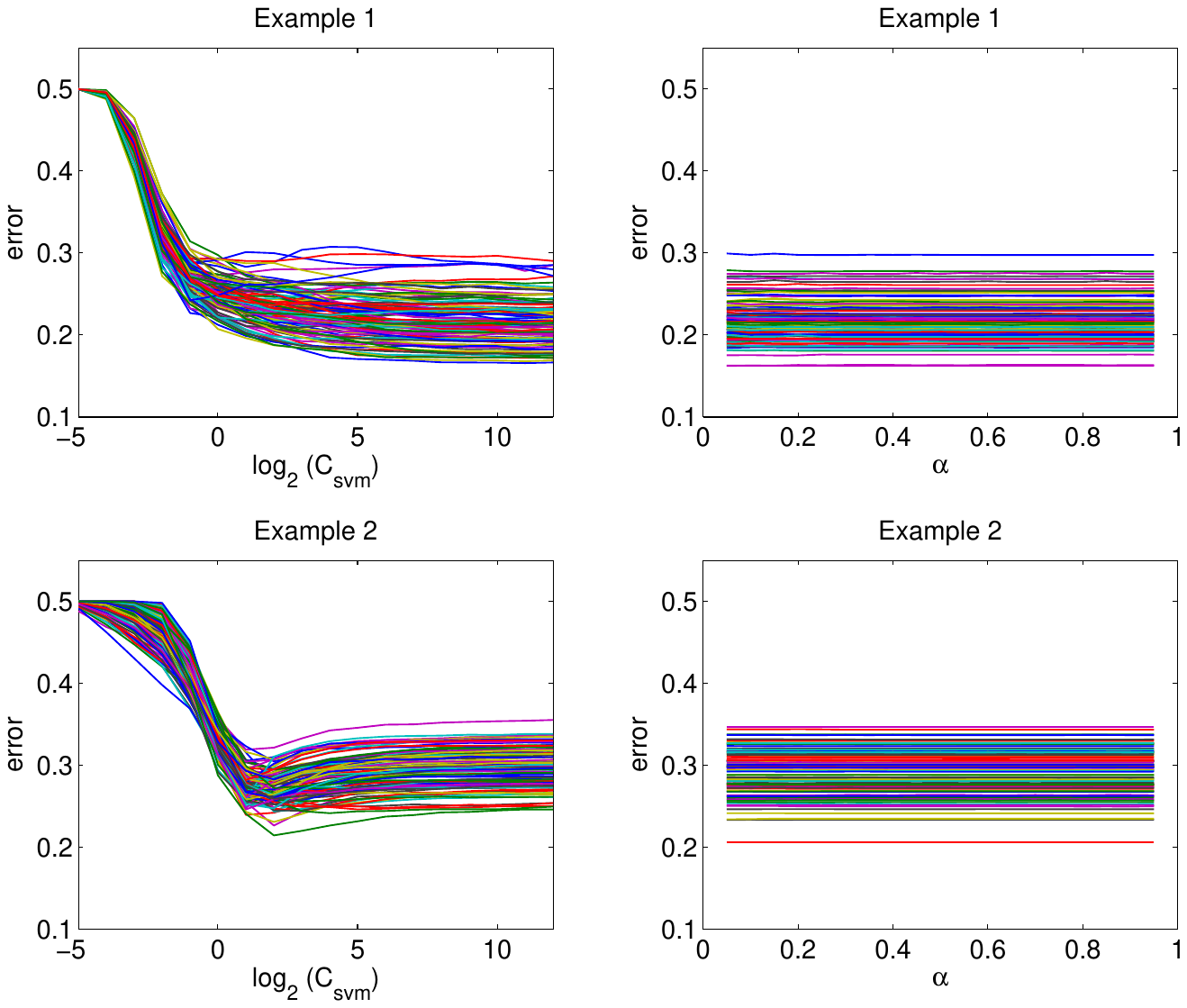}\vspace{-5ex}
	\caption{\small Left panels: Test errors of DWSVM applied to Example 1 and Example 2 (with $d=300$ and $\alpha = 0.5$) for different values of $C_{svm}$ over 100 runs. The plots show that for Example 1, any $C_{svm}$ greater than about $2^5$ will lead to similar classification performance, while for Example 2,  $C_{svm}$ around $2^1$ to $2^2$ are the best, although the performance of such parameter choice is not very different from those whose has even greater $C_{svm}$ values. Right panels: Test errors of DWSVM applied to Example 1 and Example 2 (with $d=300$ and $C_{svm}=100$) for different values of $\alpha$ over 100 runs. The performance does not depend on the choice of the $\alpha$ value very much.}
	\label{fig:sensitivity}
\end{figure}

We applied DWSVM to 100 simulations from the simulated examples defined above (Example 1 and Example 2) respectively, using the following schedules,
\begin{itemize}
	\item for fixed $\alpha=0.5$ and various values of $C_{svm}=2^{-5},2^{-4}\dots,2^{11},2^{12}$;
	\item for fixed $C_{svm}=100$ and various values of $\alpha = 0.05,1,0.1,0.15,\dots,0.95$.
\end{itemize}

Figure \ref{fig:sensitivity} reports the results. In the right panel, we show the classification error for a test data set for different values of $\alpha$. It is very clear that in these settings (where $C_{svm}=100$), the performance of DWSVM does not depend on the value of $\alpha$ as all the curves appear horizontal straight lines. It may be too early to conclude that the performance of DWSVM is independent of $\alpha$ from this observation, since it could be due to the fact that $C_{svm}=100$ happens to be a reasonably good parameter (see the discussion below). But it does suggest that the performance is influenced less by the $\alpha$ parameter than by the other parameters.

In the left panel, we do the same thing for difference values of $C_{svm}$ given $\alpha=0.5$. A similar message can be obtained, although on a restrictive condition: For Example 1, the curves appear to be flat when $C_{svm}>2^5$. Thus any value that falls into this range should work reasonable well. For Example 2, it can be seen that the optimal $C_{svm}$ is around $2^1$ and $2^2$. However, even their performance is not significantly better than those with greater $C_{svm}$. Overall, it seems that as long as the value of $C_{svm}$ is not too small, the classification performance would be close to the optimality. This is the reason why we fix the value of $\alpha$ and $C_{svm}$ to be 0.5 and 100 respectively in our comparison study conducted in the previous section. The user are free to grid search the values of $C_{svm}$ and $\alpha$  if he/she wishes so, although it seems that the effort for the latter is not worthwhile.

\section{Real application}\label{sec:real}
In this section, we compare DWSVM with the competing classifiers by applying them to the Golub data set \citep{Golub1999Molecular}. This gene expression data has 3051 genes and 38 tumor mRNA samples from the leukemia microarray study of \citet{Golub1999Molecular}. Pre-processing was done as described in \citet{Dudoit2002Comparison}.
\begin{figure}[!ht]\vspace{-2ex}
	\centering
		\includegraphics[width=\linewidth]{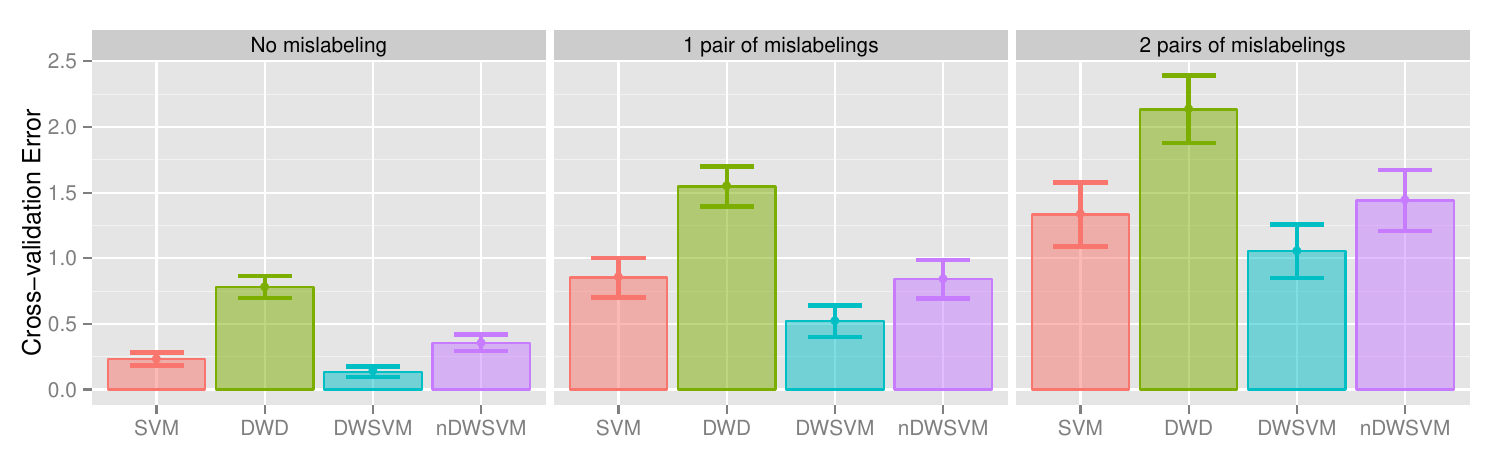}\vspace{-2ex}
	\caption{\small Cross-validated number of misclassified observations for SVM, DWD, DWSVM and nDWSVM for the original Golub data set, the Golub data with a pair of mislabeled observations, and the data with 2 pairs of mislabeled observations. For the original data, both the SVM and the DWSVM methods have CV error almost 0, with DWSVM being a little better. When there are mislabeled observations, the advantage of DWSVM becomes more obvious: it can be seen that DWSVM has the smallest CV errors while nDWSVM is on a par with SVM. The DWD classifier is always worse than the others in terms of the classification performance.}
	\label{fig:golub}
\end{figure}

As there are 11 and 27 observations from both classes, we expect the SVM and the DWSVM classifiers will give better result than DWD because the latter is subject to the imbalanced sample size. Moreover, because the dimension is much higher than the sample size, we expect severe overfitting in this data. We apply SVM, DWD, DWSVM and nDWSVM to the data set and use 3-fold cross validation to find the best $C_{svm}$ tuning parameter value. The $C_{dwd}$ and $\alpha$ values are fixed. In the left panel of Figure \ref{fig:golub}, we report the average cross-validated (CV) number of misclassfied observations and the standard error over 100 random foldings. Both SVM and DWSVM give very good result (CV error almost zero), although the DWSVM method is a little better. The nDWSVM error is almost twice that of the SVM and the DWD error is almost four times.

In order to see the extend to which our DWSVM avoids overfitting, we perturb the original data set as follows. We randomly switch the class labels of $k$ pairs of observations ($k$ observations from each class) ($k=1,2$). Then we conduct parameter tuning (via cross-validation) and training based on the perturbed data. Then, we calculate the cross-validated error for the resulting classifier: we use two folds (2/3) of the perturbed data to training a classifier, and evaluate the number of misclassified observation for the rest fold using the true class labels (the label before perturbation). Because we randomly add in noise into such settings, the CV errors increases. However, a classifier which is subject to overfitting would have a greater CV error in this setting. In the middle and the right panels of Figure \ref{fig:golub}, we report the CV error for the perturbed data where one pair and two pairs of data vectors are mislabeled respectively. As we can see, although all classifiers perform worse here than for the original data, the DWSVM classifier gives the lowest CV errors for the perturbed data. Even the performance of the two-step nDWSVM is on the par with SVM. The performance of DWD is always the worse in all three setting because of the imbalanced data issue.

\section{Theoretical properties}\label{sec:theory}
We will show some theoretical properties of DWSVM in three different favors. First, we derive the Fisher consistency of the DWSVM loss function. Note that the loss function of DWSVM is not a typical large-margin loss function. Second, we derive the asymptotic normality of the DWSVM coefficient vector. Third, we show that the intercept of DWSVM does not diverge, even in an extremely imbalanced setting.
\subsection{Fisher consistency}\label{sec:fisher}
The DWSVM method can be estimated from equations (\ref{eq:dwsvm_simple1})--(\ref{eq:dwsvm_simple2}). Thus the underlying loss function as be written as
$L(yf(\xv),yf_0(\xv)) = \alpha V_{C_{dwd}}(yf_0(\xv))+(1-\alpha)H_{C_{svm}}(yf(\xv))$. Because there are two functions involved, the underlying loss function is not a traditional margin-based loss function which involves only one function, such as that considered in \citet{Lin2004note}. Moreover, the two hyperplanes implied by $f$ and $f_0$ in our methods are parallel to each other. In general cases (beyond linear functions), this can be interpreted as the difference of these two functions is a constant, \textit{i.e.}, $f(\xv)-f_0(\xv)$ is independent of $\xv$. Theorem \ref{thm:fisher} below shows the Fisher consistency of the DWSVM loss function.
\begin{theorem}\label{thm:fisher}
For any given $C_{svm},C_{dwd}>0$ and $\alpha\in[0,1)$, if $\E [L\left\{Yf(\Xv),Yf_0(\Xv)\right\}]$ has a global minimizer $(f^*(\xv),f_0^*(\xv))$ subject to $f(\xv)-f_0(\xv)$ is a constant, then $\sgn[f^*(\xv)]=\sgn[q(\xv)-1/2]$, where $q(\xv)\equiv \Pr(Y=+1\mid \Xv=\xv)$.
\end{theorem}
Fisher consistency of the DWSVM loss function ensures that the sign of the minimizer of the expected loss function (subject to the parallel condition) coincides with the Bayes rule. 

\subsection{Asymptotic normality}\label{sec:normality}
\citet{Koo2008Bahadur} has studied the asymptotic normality of the coefficient vector for the SVM classifier. We follow the same direction and prove the corresponding results for the DWSVM classifier.

For ease of presentation of the theorem, we let $\omegav_+$ denote the augmented parameter vector $(\beta_0,\beta,\omegav^T)^T\in\R^{d+2}$, $\xv_+$, $\xv_\dagger$ and $\xv_{\ddagger}$ the augmented data vectors $(0,1,\xv^T)^T\in\R^{d+2}$, $(1,0,\xv^T)^T\in\R^{d+2}$ and $(1,1,\xv^T)^T\in\R^{d+2}$.  Consequently, the main discriminant function $f(\xv;\omegav_+)\equiv {\xv_+}^T \omegav_+=\xv^T\omegav+\beta$, and the axillary discriminant function $f_0(\xv;\omegav_+)\equiv {\xv_\dagger}^T \omegav^+=\xv^T\omegav+\beta_0$. 

We cast DWSVM to an optimization problem with an unconstrained objective function.
\begin{align}
 q_{\lambda,n}(\omegav_+) &\equiv \frac{1}{n} \sum_{i=1}^n L(\xv_i,y_i,\omegav_+)+\frac{\lambda}{2}\|\omegav\|^2\\
	&= \frac{1}{n} \sum_{i=1}^n \left\{\alpha V_{C_{d}}(y_i f_0(\xv_i;\omegav_+))+(1-\alpha)H_{C_{s}}(y_i f(\xv_i;\omegav_+))\right\}+\frac{\lambda}{2}\|\omegav\|^2\label{obj}
\end{align}
The solution to the optimization problem can be scaled by the norm of $\omegav$ so as to make it have unit norm.

The population version of (\ref{obj}) without the penalty term is defined as 
\begin{align*}
	Q(\omegav_+)\equiv \E \left\{\alpha V_{C_{d}}(Y f_0(\Xv;\omegav_+))+(1-\alpha)H_{C_{s}}(Y f(\Xv;\omegav_+))\right\}, 
\end{align*}whose minimizer is defined as $\ds\omegav_+^*\equiv\argmin_{\omegav_+}Q(\omegav_+)$.

For easy presentation, let
\begin{align*}
	g(\xv,y,\omegav_+)&\equiv \alpha \left(-\Ind{yf_0(\xv;\omegav_+)\leq 1/\sqrt{C_d}}C_d-\Ind{yf_0(\xv;\omegav_+)> 1/\sqrt{C_d}}1/[y f_0(\xv;\omegav_+)]^2\right),\\
	h(\xv,y,\omegav_+)&\equiv (1-\alpha)\left(-\Ind{y f(\xv;\omegav_+)\leq 1/\sqrt{C_s}}C_s\right),\\
	v(\xv,y,\omegav_+)&\equiv \alpha\left(\Ind{yf_0(\xv;\omegav_+)> 1/\sqrt{C_d}}1/[y f_0(\xv;\omegav_+)]^3\right),\\
	w(\xv,y,\omegav_+)&\equiv (1-\alpha)\delta\left(1/\sqrt{C_s}-yf(\xv;\omegav_+)\right)C_s,
\end{align*}
where $\delta(\cdot)$ denotes the Dirac delta function. Furthermore, let
\begin{align*}
	S(\omegav_+)&\equiv \E\left\{g(\Xv,Y,\omegav_+)Y\Xv_\dagger+h(\Xv,Y,\omegav_+)Y\Xv_+\right\}\mb{ and}\\
	U(\omegav_+)&\equiv \E\left\{v(\Xv,Y,\omegav_+)\Xv_\dagger\Xv_\dagger^T+w(\xv,y,\omegav_+)\Xv_+\Xv_+^T\right\}.
\end{align*}

Let $ \Omega(\Xv_i,Y_i,\omegav_+^*) = \diag\{g(\Xv_i,Y_i,\omegav_+^*),h(\Xv_i,Y_i,\omegav_+^*),[g(\Xv_i,Y_i,\omegav_+^*)+h(\Xv_i,Y_i,\omegav_+^*)]\Id_d\}$, where $\Id_d$ is $d\times d$ identity matrix.

Then, define
\begin{align*}
		T_n\equiv& \sum_{i=1}^n\bigg\{ g(\Xv_i,Y_i,\omegav_+^*)Y_i(\Xv_i)_\dagger+h(\Xv_i,Y_i,\omegav_+^*)Y_i(\Xv_i)_+\bigg\},\\
		=&\sum_{i=1}^nY_i\bigg\{ \Omega(\Xv_i,Y_i,\omegav_+^*)(\Xv_i)_\ddagger\bigg\}.
\end{align*}

Lastly, define $G(\omegav_+^*)\equiv\E\left[(\Xv_i)_\ddagger \Omega^2(\Xv_i,Y_i,\omegav_+^*){(\Xv_i)_\ddagger}^T\right]$.

Some regularity conditions are needed. We state the conditions in the appendix. Note that conditions (A1), (A2) and (A4) are the same as in \citet{Koo2008Bahadur}.  Our new (A3) is tailored for DWSVM and incorporates the DWD component. In particular, (A1) ensures that $U(\omegav^+)$ is well-defined and is continuous in $\omegav_+$ while (A1) and (A2) ensure that the minimizer $\omegav_+^*$ exists. (A3) is a sufficient condition to that $\omegav_+^*$ is not zero. (A4) guarantees the positive-definiteness of $U(\omegav_+)$ around $\omegav_+^*$.

Under these regularity conditions, we obtain a Bahadur representation of $\wh{\omegav_{\lambda,n}}_+$ in Theorem \ref{bahadur}, the asymptotic normality in Theorem \ref{normality}, and consequently, the asymptotic normality of the discriminant function $f(\xv;\wh{\omegav_{\lambda,n}}_+)$ at $\xv$ in Corollary \ref{normality2}.

\begin{theorem}\label{bahadur}
Suppose that (A1)--(A4) are met. For $\lambda=o(n^{-1/2})$, we have 
$$\sqrt{n}(\wh{\omegav_{\lambda,n}}_+-\omegav_+^*)=-\frac{1}{\sqrt{n}}U(\omegav_+^*)^{-1}T_n+o_P(1).$$
\end{theorem}

\begin{theorem}\label{normality}
Suppose that (A1)--(A4) are met. For $\lambda=o(n^{-1/2})$, we have 
$$\sqrt{n}(\wh{\omegav_{\lambda,n}}_+-\omegav_+^*)=N\left(\0v,U(\omegav_+^*)^{-1} G(\omegav_+^*) U(\omegav_+^*)^{-1}\right)$$
\end{theorem}

This will lead to the following corollary. 
\begin{corollary}\label{normality2}
Under the same conditions as in Theorem \ref{normality}, for $\lambda=o(n^{-1/2})$ and any $\xv\in\R^d$,
$$\sqrt{n}\left(f(\xv,\wh{\omegav_{\lambda,n}}_+)-f(\xv,\omegav_+^*)\right)\stackrel{d}{\rightarrow}N\left(\0v,\xv_+^TU(\omegav_+^*)^{-1} G(\omegav_+^*) U(\omegav_+^*)^{-1}\xv_+\right)$$
\end{corollary}

\subsection{Extremely imbalanced data}\label{sec:imbalance}
\citet{Owen2007Infinitely} discussed the behavior of the intercept term in the logistic regression when the sample size of one class is extremely large while that of the other class is fixed. Moreover, \citet{Qiao13Flexible} also showed that the intercept term of DWD diverges. In this subsection, we prove that the intercept term for the DWSVM classifier does not diverge. Without loss of generality, we assume that $n_-\gg n_+$, \ie, the negative class is the majority class.

\begin{lemma}\label{thm:svm_imbalanced}
Suppose that the negative majority class is sampled from a distribution with compact support $\Sc$. Then the intercept term $\beta$ in SVM does not diverge to negative infinity when $n_-\rightarrow \infty$.
\end{lemma}

\begin{corollary}\label{thm:dwsvm_imbalanced}
Suppose that the negative majority class is sampled from a distribution with compact support $\Sc$. Then the intercept term $\beta$ in DWSVM does not diverge to negative infinity when $n_-\rightarrow \infty$.
\end{corollary}

The assumption of compact support $\Sc$ is essential here, but it is fairly weak and is true in many real applications. Note that this result does not ensure that the sensitivity issue is completely overcome by SVM or DWSVM. Instead, it suggests that in the $n_-\rightarrow\infty$ asymptotics, the impact of the imbalanced sample size is limited to some extent.

\section{Conclusion}\label{sec:conclude}
Both SVM and DWD are subject to certain disadvantages and enjoy certain advantages. The DWSVM combines the merits of both methods by creatively deploying an axillary intercept term. We have shown standard asymptotic results for the DWSVM classifier. The simulations and real data application establish the superiority of the DWSVM method over SVM and DWD in some situations. In particular, the DWSVM method can lead to a discriminant direction vector that, like the DWD direction, preserve important features of the data set. More importantly, the DWSVM also performs very well in terms of classification. As a bottom line, its performance is just as good as the SVM. In special settings such as the perturbed data, we have demonstrated that DWSVM can overcome overfitting and is more robust against perturbation/mislabeling of the data.

We have shown some asymptotic properties of DWSVM in this paper. More work can be done to investigate its statistical properties, for example, in the line of \citet{blanchard2008statistical}.

An instant extension of the DWSVM classifier is multiclass classification. For example, for a multiclass classification problem with $K$ classes, the following optimization problem accomplishes such an extension.
\begin{align*}
\argmin_{\framebox{$\omegav_j,\beta_j$}\beta_{j0},\xiv,\etav} &\sum_{i=1}^n \sum_{y_i=j,~k\neq j}\left\{\alpha\left(\frac{1}{r_{jk}^i}+C_{dwd}\cdot \eta_{jk}^i\right)+(1-\alpha)\xi_{jk}^i\right\},\\
\mb{s.t.}\quad 	& r_{jk}^i=y_i\{\xv_i^T(\omegav_j-\omegav_k)+(\beta_{j0}-\beta_{k0})\}+\eta_{jk}^i,~r_{jk}^i\geq 0 \mb{ and } \eta_{jk}^i\geq 0,\\
								& C_{svm}y_i \{\xv_i^T(\omegav_j-\omegav_k)+(\beta_{j}-\beta_{k})\}+\xi_{jk}^i\geq \sqrt{C_{svm}},~\xi_{jk}^i\geq 0,\\
								&\sum_{j=1}^K\|\omegav_j\|^2\leq 1,\\
								&\sum_{j=1}^K\omegav_j = \0v,~\sum_{j=1}^K\beta_j=0,~\sum_{j=1}^K\beta_{j0}=0.
\end{align*}
Other extensions such as kernel DWSVM or sparse DWSVM are also readily in order.

In summary, DWSVM integrates the merits of classical classification methods. Its numerical performance is very good and it is theoretically justified. These show evidence that it is a very promising linear learner which has great potential in many applications.

Future work will also concentrate on developing more efficient implementation of DWSVM.

\section*{Acknowledgment}
The first author's work was partially supported by Binghamton University Harpur College Dean's New Faculty Start-up Funds and a collaboration grant from the Simons Foundation (\#246649 to Xingye Qiao). Both authors thank the Statistical and Applied Mathematical Sciences Institute (SAMSI) for their generous support where both authors have spent considerable amount of time when writing this article.

\section*{Appendices}\label{sec:appendix}
\subsection*{Proof of Theorem \ref{thm:fisher}}
For any $\xv$, denote $q(\xv) = \Pr(Y=+1|\Xv=\xv)$. The conditional risk is
\begin{align*}
	R(f,f_0)&\equiv E[L\left\{Yf(\Xv),Yf_0(\Xv)\right\}\mid \Xv=\xv]\\
		  &=\left\{\alpha V_{C_{dwd}}(f_0)+(1-\alpha) H_{C_{svm}}(f)\right\}q(\xv)\\
		  &\quad\quad+\left\{\alpha V_{C_{dwd}}(-f_0)+(1-\alpha) H_{C_{svm}}(-f)\right\}\{1-q(\xv)\},
\end{align*} where for simplicity we write $f(\xv)$ and $f_0(\xv)$ as $f$ and $f_0$.

For the global minimizer $(f^*,f_0^*)$, since $f^*-f_0^*=\Delta^*$ is independent of $\xv$, we can consider another feasible (but not optimal) solution $(-f^*,-f^*-\Delta^*)$. Due to the optimality of $(f^*,f_0^*)=(f^*,f^*-\Delta^*)$, we can show that
\begin{align*} 
		0\geq &R(f^*,f^*-\Delta^*)-R(-f^*,-f^*-\Delta^*)\\
	= &\{2q(\xv)-1\}\left[\{\alpha V_{C_{dwd}}(f^*-\Delta^*)+(1-\alpha)H_{C_{svm}}(f^*)\}-\{\alpha V_{C_{dwd}}(-f^*-\Delta^*)+(1-\alpha)H_{C_{svm}}(-f^*)\}\right]\\
	= &\{2q(\xv)-1\}\left[\alpha\{V_{C_{dwd}}(f^*-\Delta^*)-V_{C_{dwd}}(-f^*-\Delta^*)\}+(1-\alpha)\{H_{C_{svm}}(f^*)-H_{C_{svm}}(-f^*)\}\right]
\end{align*}
Thus if $q(\xv)>1/2$, then $$\alpha\{V_{C_{dwd}}(f^*-\Delta^*)-V_{C_{dwd}}(-f^*-\Delta^*)\}+(1-\alpha)\{H_{C_{svm}}(f^*)-H_{C_{svm}}(-f^*)\}\leq 0.$$ Because $V_{C_{dwd}}(\cdot)$ is strictly decreasing everywhere, and $H_{C_{svm}}(\cdot)$ is strictly decreasing around 0, we have that $V_{C_{dwd}}(f^*-\Delta^*)-V_{C_{dwd}}(-f^*-\Delta^*)$ and $H_{C_{svm}}(f^*)-H_{C_{svm}}(-f^*)$ have the same sign, and hence $f^*\geq 0$. By a similar argument, if $q(\xv)<1/2$, then $f^*\leq 0$. Lastly, it is easy to show that $f^*\neq 0$. Hence we have $\sgn(f^*) = \sgn(q(\xv)-1/2)$.

\subsection*{Regularity conditions}
We state the regularity conditions for the asymptotics below. We use $C_1$, $C_2$, \dots to denote positive constants independent of $n$.

\begin{description}
	\item[A1] The densities $p_+$ and $p_-$ are continuous and have finite second moments.
	\item[A2] There exists $B(\xv_0,\delta_0)$, a ball centered at $\xv_0$ with radius $\delta_0$ such that $p_1(\xv)>C_1$ and $p_2(\xv)>C_1$ for every $\xv\in B(\xv_0,\delta_0)$.
	\item[A3] For some $1\leq l\leq d$,
	$$\E\left(\Ind{X_l\geq F_-^L}X\mid Y=-1\right)<\E\left(\Ind{X_l\leq F_+^U}X\mid Y=+1\right)$$
	or
	$$\E\left(\Ind{X_l\leq F_-^U}X\mid Y=-1\right)>\E\left(\Ind{X_l\geq F_+^L}X\mid Y=+1\right),$$
	where $F_+^L$ and$F_-^L$ ($F_+^U$ and $F_-^U$, respectively) are the lower bounds (upper bounds, respectively) for the positive and negative classes. They are defined as
\begin{align*}
		\Pr\left(X_l\geq F_+^L \mid Y=+1 \right) &= \min\left(1,\frac{\pi_+\{\alpha{C_d}+(1-\alpha){C_s}\}}{\pi_-(1-\alpha){C_s}}\right),\\
		\Pr\left(X_l\geq F_-^L \mid Y=+1 \right) &= \min\left(1,\frac{\pi_-\{\alpha{C_d}+(1-\alpha){C_s}\}}{\pi_+(1-\alpha){C_s}}\right),\\
		\Pr\left(X_l\leq F_+^U \mid Y=+1 \right) &=  \min\left(1,\frac{\pi_+(1-\alpha){C_s}}{\pi_-\{\alpha{C_d}+(1-\alpha){C_s}\}}\right),\\		\Pr\left(X_l\leq F_-^U \mid Y=+1 \right) &=\min\left(1,\frac{\pi_-(1-\alpha){C_s}}{\pi_+\{\alpha{C_d}+(1-\alpha){C_s}\}}\right).
\end{align*}
		\item[A4] For an orthogonal transformation $A_l$ that maps $\omegav^*/\|\omegav^*\|$ to the $l$th unit basis vector $e_l$ for some $1\leq l\leq d$, there exist rectangles
		$$\Dc^+ = \set{\xv\in M^+: l_s\leq (A_l\xv)_s\leq v_s\mb{ with }l_s<v_s\mb{ for }s\neq l}$$
and
	$$\Dc^- = \set{\xv\in M^-: l_s\leq (A_l\xv)_s\leq v_s\mb{ with }l_s<v_s\mb{ for }s\neq l}$$
	such that $p_+(\xv)\geq C_2>0$ on $\Dc^+$ and $p_-(\xv)\geq C_3>0$ on $\Dc^-$, where $M^+\equiv\set{\xv: \xv^T\omegav^*+\beta=1/\sqrt{C_s}}$ and $M^-\equiv\set{\xv: \xv^T\omegav^*+\beta=-1/\sqrt{C_s}}$.
\end{description}

\subsection*{Proof of Theorems \ref{bahadur} and \ref{normality} and Corollary \ref{normality2}}
For fixed $\thetav\in\R^{d+2}$, define 
\begin{align*}
	\Lambda_n(\thetav)&\equiv n\left\{q_{\lambda,n}(\omegav_+^*+\thetav/\sqrt{n})-q_{\lambda,n}(\omegav_+^*)\right\},\mb{ and}\\
	\Gamma_n(\thetav)&\equiv \E\Lambda_n(\thetav).
\end{align*}

Observe that 
$$\Gamma_n(\thetav) = n\left\{Q(\omegav_+^*+\thetav/\sqrt{n})-Q(\omegav_+^*)\right\}+\frac{\lambda}{2}\left(\|\thetav_{3:(d+2)}\|^2+2\sqrt{n}{\omegav^*}^T\thetav_{3:(d+2)}\right)$$
By Taylor series expansion of $Q$ around $\omegav_+^*$, we obtain, for some $0<t<1$,
$$\Gamma_n(\thetav) = \frac{1}{2}\thetav^TU\left(\omegav_+^*+(t/\sqrt{n})\thetav\right)\thetav
+\frac{\lambda}{2}\left(\|\thetav_{3:(d+2)}\|^2+2\sqrt{n}{\omegav^*}^T\thetav_{3:(d+2)}\right).$$
Because $U(\omegav_+)$ is continuous in $\omegav_+$, due to condition (A1), we have 
$$\frac{1}{2}\thetav^TU\left(\omegav_+^*+(t/\sqrt{n})\thetav\right)\thetav = \frac{1}{2}\thetav^TU\left(\omegav_+^*\right)\thetav+o(1).$$This, combined with $\lambda=o(n^{-1/2})$, results in
$$\Gamma_n(\thetav) = \frac{1}{2}\thetav^TU\left(\omegav_+^*\right)\thetav
+o(1).$$

Now, observe that 
$\E T_n = nS(\omegav_+^*) = \0v$ and $\E (T_nT_n^T) = \sum_{i=1}^n \E\left[(\Xv_i)_\ddagger
\Omega^2(\Xv_i,Y_i,\omegav_+^*){(\Xv_i)_\ddagger}^T\right]=nG(\omegav_+^*)$. Hence, $\frac{1}{\sqrt{n}}T_n$ follows $N\left(0,G(\omegav_+^*)\right)$ asymptotically by central limit theorem.

Next, we define
$$R_{i,n}(\thetav) \equiv L_{i,n}(\omegav_+^*+\thetav/\sqrt{n})-L_{i,n}(\omegav_+^*)-\left(\frac{\partial L_{i,n}}{\partial \omegav_+}(\omegav_+)\bigg|_{\omegav_+=\omegav_+^*}\right)^T\thetav/\sqrt{n},$$
where $L_{i,n}(\omegav_+)\equiv \alpha V_{C_{d}}(Y_i(\Xv_i)_\dagger^T\omegav_+)+(1-\alpha)H_{C_{s}}(Y_i(\Xv_i)_+^T\omegav_+)$.

We continue by splitting $R_{i,n}$ to two parts $R_{i,n}=\alpha R^d_{i,n}+(1-\alpha)R^s_{i,n}$, where the first term concerns the DWD component and the second term concerns the SVM component.

For the DWD component,
$$R_{i,n}^d(\thetav) \equiv V(Y_i(\Xv_i)_\dagger^T(\omegav_+^*+\thetav/\sqrt{n}))-V(Y_i(\Xv_i)_\dagger^T\omegav_+^*)-\left(\frac{\partial V}{\partial \omegav_+}(\omegav_+)\bigg|_{\omegav_+=\omegav_+^*}\right)^T\thetav/\sqrt{n}$$
Because the DWD loss $V$ has first order continuous derivative, $R_{i,n}^d(\thetav)=O(n^{-1})$.

For the SVM component, 
$$R_{i,n}^s(\thetav) \equiv H[Y_i(\Xv_i)_+^T(\omegav_+^*+\thetav/\sqrt{n})]-H[Y_i(\Xv_i)_+^T\omegav_+^*]+\sqrt{C_s}\Ind{Y_i(\Xv_i)_+^T\omegav_+^*<1/\sqrt{C_s}}Y_i(\Xv_i)_+^T\thetav/\sqrt{n}.$$

Following the argument by \citet{Koo2008Bahadur} and combining the fact that $R_{i,n}^d(\thetav)=O(n^{-1})$, we can show that $\sum_{i=1}^n\E\left(|R_{i,n}(\thetav)-\E R_{i,n}(\thetav)|^2\right)\rightarrow 0$, as $n\rightarrow 0$

We note that $\Lambda_n(\thetav) = \Gamma_n(\thetav)+T_n^T\thetav/\sqrt{n}+\sum_{i=1}^n \left(R_{i,n}(\thetav)-\E R_{i,n}(\thetav)\right).$ Thus
\begin{align*}
	\Lambda_n(\thetav) &= \frac{1}{2}\thetav^TU\left(\omegav_+^*\right)\thetav+T_n^T\thetav/\sqrt{n}+o_P(1).
\end{align*}

By the Convexity Lemma in \citet{Pollard1991Asymptotics}, we have for any fixed $\theta$,
\begin{align*}
	\Lambda_n(\thetav) &= \frac{1}{2}(\thetav-\zetav_n)^TU\left(\omegav_+^*\right)(\thetav-\zetav_n)+\frac{1}{2}\zetav_n^TU\left(\omegav_+^*\right)\zetav_n+r_n(\theta),
\end{align*}where $\zetav_n\equiv -U(\omegav_+^*)^{-1}T_n/\sqrt{n}$, and for each compact set $K\in\R^d$,
$$\sup_{\thetav\in K} |r_n(\thetav)|\stackrel{p}{\rightarrow} 0.$$

We then follow the argument in \citet{Koo2008Bahadur} and have for each $\varepsilon>0$ and $\wh\thetav_{\lambda,n}=\sqrt{n}(\wh{\omegav_{\lambda,n}}_+-\omegav_+^*)$, 
$$\Pr \left(\|\wh\thetav_{\lambda,n}-\zetav_n\|>\varepsilon\right)\stackrel{p}{\rightarrow} 0,$$ which completes the proof.\hfill\qed

\subsection*{Proof of Lemma \ref{thm:svm_imbalanced}}
We prove the result for the simpler and more intuitive case of $d=1$. In this case $\omegav\in\R$ does not need to be optimized. We can simply assume that $\omegav = 1$. Moreover, we can consider the worst case scenario where $n_+=1$. This is the worse case because this represents the most imbalanced sample sizes. We let $x_0$ denote the sole data vector in the positive minority class

Since the negative class is extremely large compared to the positive, we can assume that the functional margin with respective to the main hyperplane $u\equiv y_0(x_0+\beta)=x_0+\beta$ for the data vectors from the positive minority class are always less than $1/\sqrt{C_s}$, that is $\beta\leq 1/\sqrt{C_s}-x_0$.

Write the objective function of SVM as 
\begin{align*}
	\Lc^s(\beta)\equiv&(\sqrt{C_s}-C_s x_0-C_s\beta)+\sum_{i=1}^n\left\{\Ind{y_i=-1}(\sqrt{C_s}+C_sx_i+C_s\beta)_+\right\}\\
	\approx&(\sqrt{C_s}-C_s x_0-C_s\beta)+n_- \E\left\{(\sqrt{C_s}+C_sX+C_s\beta)_+\mid Y=-1\right\}
\end{align*}
Note that
\begin{align*}
	\frac{\partial \Lc^s}{\partial \beta}(\beta)\equiv&-C_s+n_-C_s \E\left[\Ind{\sqrt{C_s}+C_sX+C_s\beta>0}\mid Y=-1\right]\\
	=&-C_s+n_-C_s \Pr\left[\sqrt{C_s}+C_sX+C_s\beta>0\mid Y=-1\right].
\end{align*}

This leads to that
\begin{align*}
	\lim_{\beta\rightarrow -\infty}\frac{\partial \Lc^s}{\partial \beta}(\beta)=&-C_s<0\\
	\frac{\partial \Lc^s}{\partial \beta}\left(-M-1/\sqrt{C_s}\right)=&-C_s+n_-C_s \Pr\left[\sqrt{C_s}+C_sX+C_s(-M-1/\sqrt{C_s})>0\mid Y=-1\right]\\
	=&-C_s+n_-C_s \Pr\left[X>M\mid Y=-1\right]=-C_s<0
\end{align*}
Thus if $1/\sqrt{C_s}-x_0\leq -M-1/\sqrt{C_s}$, then 
\begin{align*}
	\frac{\partial \Lc^s}{\partial \beta}\left(1/\sqrt{C_s}-x_0\right)=&-C_s+n_-C_s \Pr\left[\sqrt{C_s}+C_sX+C_s(1/\sqrt{C_s}-x_0)>0\mid Y=-1\right]\\
	=&-C_s+n_-C_s \Pr\left[X>x_0-2/\sqrt{C_s}\mid Y=-1\right]\\
	=&-C_s<0,
\end{align*}
and $\beta = 1/\sqrt{C_s}-x_0$ is the minimizer of $\Lc^s$. On the other hand, if  $1/\sqrt{C_s}-x_0> -M-1/\sqrt{C_s}$, then the minimizer $\beta^*$ will be greater than $-M-1/\sqrt{C_s}$ but less than or equal to $1/\sqrt{C_s}-x_0$. This means that the intercept term $\beta$ in SVM does not diverge to $-\infty$.\hfill\qed

\bibliographystyle{asa}
\bibliography{DWSVM_bib}
\end{document}